\documentclass{article}

\usepackage{microtype}
\usepackage{graphicx}
\usepackage{subfigure}
\usepackage{booktabs} 
\usepackage{placeins}
\usepackage{hyperref}

\usepackage{url}

\usepackage{tabularx}
\usepackage{newfloat}
\usepackage{listings}

\usepackage{subcaption}
\usepackage{multirow}

\usepackage{xcolor}
\usepackage[ruled, vlined]{algorithm2e}

\usepackage{algorithmic}

\usepackage{amsmath,amsfonts,bm}

\def\eqref#1{equation~\ref{#1}}

\def\1{\bm{1}}

\DeclareMathAlphabet{\mathsfit}{\encodingdefault}{\sfdefault}{m}{sl}
\SetMathAlphabet{\mathsfit}{bold}{\encodingdefault}{\sfdefault}{bx}{n}

\usepackage{amsmath}
\usepackage{amssymb}
\usepackage{mathtools}
\usepackage{amsthm}
\usepackage{natbib}
\usepackage[capitalize,noabbrev]{cleveref}

\newcommand{\sysname}{ODD}

\usepackage[preprint]{neurips_2025}

\usepackage[utf8]{inputenc} 
\usepackage[T1]{fontenc} 
\usepackage{hyperref} 
\usepackage{url}
\usepackage{booktabs} 
\usepackage{amsfonts} 
\usepackage{nicefrac} 
\usepackage{microtype}
\usepackage{xcolor}

\usepackage[textsize=tiny]{todonotes}

\title{Free Lunch for Pass@$k$? Low Cost Diverse Sampling for Diffusion Language Models}

\author{%
  Sean Lamont\textsuperscript{1,2}, Christian Walder\textsuperscript{3}, Paul Montague\textsuperscript{2}, Amir Dezfouli\textsuperscript{4},  Michael Norrish\textsuperscript{1} \\
  \textsuperscript{1}Australian National University\\
  \textsuperscript{2}Defence Science and Technology Group \\
  \textsuperscript{3}Google DeepMind\\
  \textsuperscript{4}BIMLOGIQ\\
  \texttt{sean.lamont@anu.edu.au}
}

\begin{document}
\maketitle

\begin{abstract}
Diverse outputs in text generation are necessary for effective exploration in complex reasoning tasks, such as code generation and mathematical problem solving. Such Pass@$k$ problems benefit from distinct candidates covering the solution space. However, traditional sampling approaches often waste computational resources on repetitive failure modes. While Diffusion Language Models have emerged as a competitive alternative to the prevailing Autoregressive paradigm, they remain susceptible to this redundancy, with independent samples frequently collapsing into similar modes. To address this, we propose a training free, low cost intervention to enhance generative diversity in Diffusion Language Models. Our approach modifies intermediate samples in a batch sequentially, where each sample is repelled from the feature space of previous samples, actively penalising redundancy. Unlike prior methods that require retraining or beam search, our strategy incurs negligible computational overhead, while ensuring that each sample contributes a unique perspective to the batch. We evaluate our method on the HumanEval and GSM8K benchmarks using the LLaDA-8B-Instruct model. Our results demonstrate significantly improved diversity and Pass@$k$ performance across various temperature settings. As a simple modification to the sampling process, our method offers an immediate, low-cost improvement for current and future Diffusion Language Models in tasks that benefit from diverse solution search. We make our code available at \url{https://github.com/sean-lamont/odd}.
\end{abstract}

\section{Introduction}
Diffusion Language Models (DLMs) such as LLaDA~\citep{nie_large_2025} have recently demonstrated performance comparable to Autoregressive (AR) models while offering distinct advantages, including faster parallel generation, superior constraint adherence~\citep{li_diffusion-lm_2022} and flexible generation controls that allow users to trade off speed and quality.
Despite these capabilities,
a persistent challenge in generation for both paradigms is \textit{mode collapse} or redundancy.
When sampling multiple solutions, standard techniques such as temperature scaling or beam search often produce highly correlated outputs.
This redundancy diminishes the utility of multiple samples in Pass@$k$ tasks such as coding or mathematics, as they result in repetitive outputs which do not effectively explore the possible solution space.

Diversity has recently become an active topic in Reinforcement Learning (RL) post-training for Language Models~\citep{tuyls_representation-based_2026, li_jointly_2025} where it is often a prerequisite for learning from sparse rewards. This is particularly critical in domains like competition mathematics, where correct solutions are rare. In such settings, a sampling strategy that collapses to a single mode may never generate a correct trajectory, even if the solution lies within the base model's capabilities. By improving diverse sampling to maximize Pass@$k$, we increase the likelihood of discovering rare successful paths, thereby providing the valuable reward signal needed to drive optimisation.
To date, the primary research focus has been on post-training interventions incorporating diversity directly into the optimisation objective~\citep{li_jointly_2025, tuyls_representation-based_2026}.
Since the initial generation forms the basis of what is selected for downstream tasks or training, effective \textit{inference-time} diversity remains an orthogonal and complementary objective.

\begin{figure}[t]
 \centering
 \includegraphics[width=1.0\linewidth]{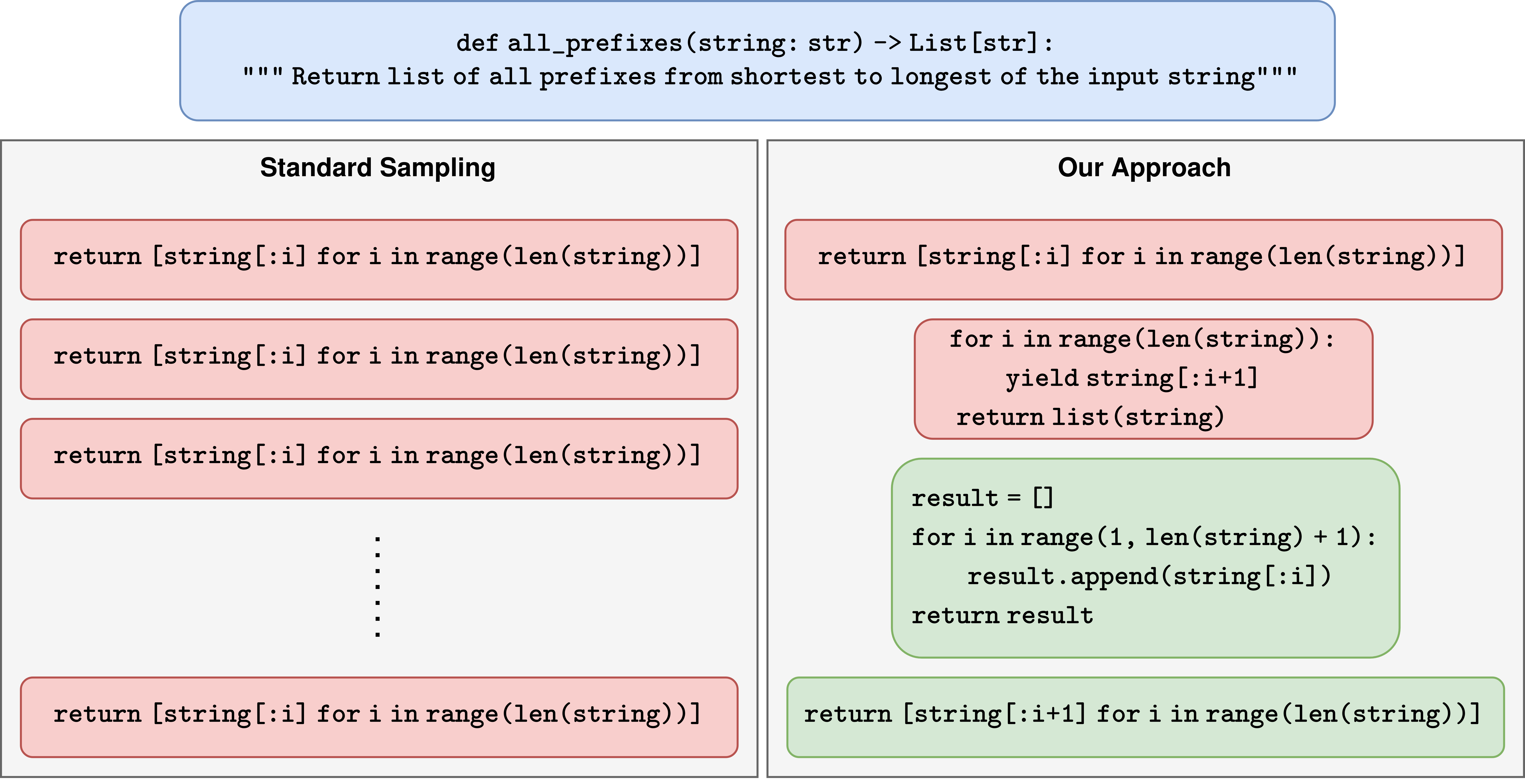}
 \caption{Example outputs from LLaDA with standard sampling (left) and our approach (right), both with a temperature of 1. For this example, the base sampler achieved a Pass@16 of 0, while our model found 3 valid solutions. Despite having the same temperature, standard sampling leads to mode collapse over an incorrect approach, whereas our method optimises to select samples distinct from those previously explored.}  
 \label{fig:example}
\end{figure}

Prior work in inference-time diversity has been restricted to AR models, focusing largely on diverse variants of beam search~\citep{vijayakumar_diverse_2016, li_simple_2016, meister_determinantal_2021, shi_semantic-guided_2025}. These approaches are restricted to AR decoding and often require training separate value models or incur significant latency penalties due to complex beam management~\citep{shi_semantic-guided_2025}.
Unlike AR models, diffusion models possess a global view of the sample at every inference step. This allows for methods which leverage both the current context and the model's predictions for the remaining tokens. 
This has been exploited in the image and protein synthesis domains, where sample trajectories are modified to optimise a batch-wide diversity term. Until now, this has not been explored in the context of text diffusion models~\citep{ferragu_diversity-guided_2024, morshed_diverseflow_2025}.
 
To address these limitations we propose \sysname{}: \textbf{O}rthogonal \textbf{D}iverse \textbf{D}iffusion, which encourages diverse (`odd') outputs to be sampled during inference. We use the intermediate states of the diffusion process to generate feature vectors, for which we apply a lightweight repulsion term during inference. As we generate a batch of $n$ samples, the logits for the $i$-th sample are actively pushed away from the subspace spanned by samples $\{1,\dots,i-1\}$. This ensures that the additional computation needed for larger batches effectively contributes to the exploration of solution space. We validate our approach on HumanEval and GSM8K, demonstrating that we can boost Pass@$k$ with minimal overhead.
To summarise our contributions:
\begin{itemize}
 \item We present a training free framework for improving generative diversity in DLMs with minimal time and space overhead. We use this to introduce a diversity improving loss based on maximising the component of the current sample orthogonal to previous samples. 
 \item Testing our approach over HumanEval and a subset of GSM8K, we demonstrate significant and consistent improvements across various temperature settings. 
 \item We open source our code. This can be used for immediate improvements to Pass@$k$ for DLMs, while our codebase allows for easy experimentation and evaluation over different feature extractors, diversity terms and benchmarks to help encourage further improvements in the area. In addition, we release all experiment logs and data for full transparency and reproducibility, available at~\url{https://sean-lamont.github.io/odd/}. 
\end{itemize}

\subsection{Related Work}
Recent works have integrated diversity directly into RL post-training~\citep{tuyls_representation-based_2026, li_jointly_2025}. By improving inference-time diversity, our method is complementary and applicable to any pre-trained diffusion model without modification. Several inference-time algorithms have been proposed for the AR domain. Diverse variants of beam search~\citep{meister_determinantal_2021, vijayakumar_diverse_2016, li_simple_2016} have been effective, though can incur significant computational overhead~\citep{shi_semantic-guided_2025}. Other strategies involve conditioning the model on diverse prompts or random seeds, but do not address redundancy over multiple attempts as they treat samples independently~\citep{nagarajan_roll_2025}.
Other work has tried to explicitly condition on past attempts to force exploration, but require previously generated output before enforcing diversity~\citep{lu_benchmarking_2025}.
Similarly, recent approaches which filter diverse sets from a larger pool are limited by the overhead of generating the initial candidates~\citep{tuyls_representation-based_2026, lamont_3d-prover_2025}. In contrast, we enforce diversity during the generation process itself, with minimal overhead. 

In the context of Diffusion Models, inference-time diversity has been explored for image and protein synthesis~\citep{morshed_diverseflow_2025, ferragu_diversity-guided_2024}.
DiverseFlow~\citep{morshed_diverseflow_2025} and similar methods~\citep{ferragu_diversity-guided_2024} modify the sampling trajectory by optimising a batch-wide diversity objective.
As part of our work we extend DiverseFlow~\citep{morshed_diverseflow_2025} to the text domain (Section~\ref{sec:results}), where we find improvements upon standard sampling. We then demonstrate significant additional improvements using our novel \sysname{} algorithm. 
For text diffusion, recent work has attempted to promote diversity by adding noise to embeddings~\citep{wu_time-annealed_2026}.
By treating samples independently, it does not effectively account for what has already been explored in a given batch.
Our approach explicitly models the dependency between samples, sequentially projecting each new sample away from the subspace spanned by previous generations. This ensures structured exploration rather than random variance.

\section{Methodology}

\begin{figure}[t]
 \centering
 \includegraphics[width=1.0\linewidth]{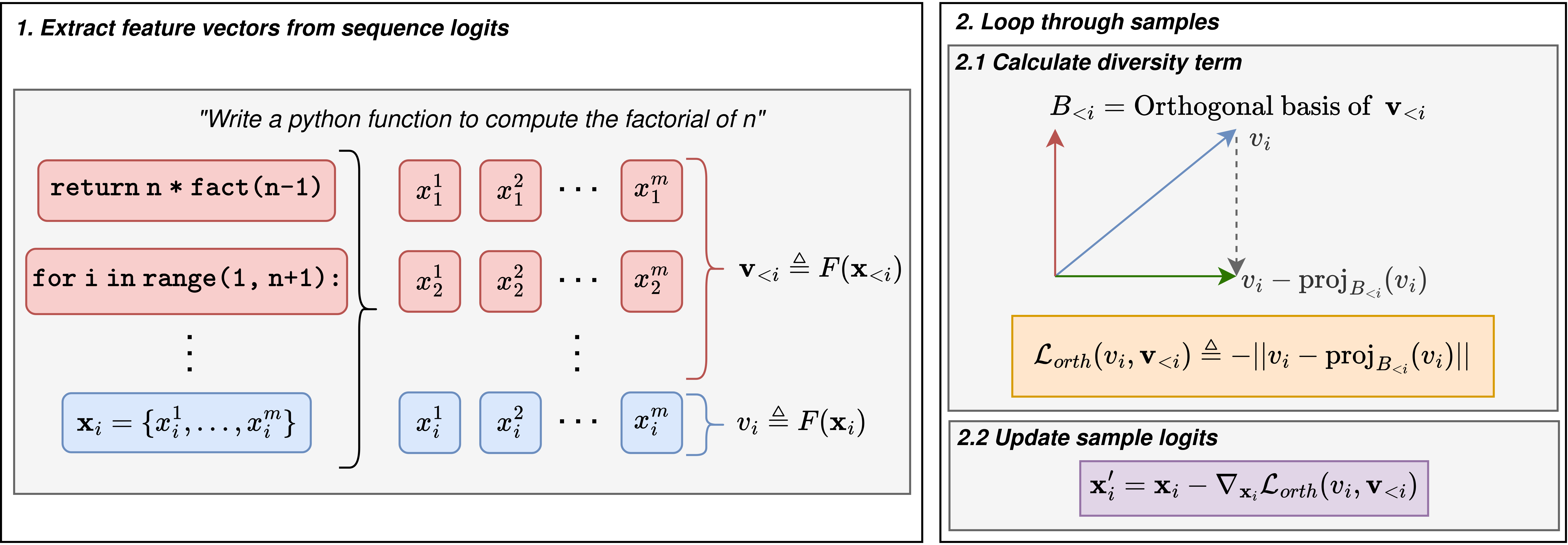}
 \caption{An overview of our proposed diverse sampling framework, using our orthogonal diversity loss. Given a full batch of $n$ samples, feature vectors $F(\mathbf{x_i})$ are extracted for the logits $x_i$ for each sample $i \in \{1, \dots, n\}$. 
 We then loop through each sample, compute an orthogonal basis for all previous sample features $B_{<i}$, and calculate the projection of the current sample features onto this basis. We then take the norm of the residual with respect to this projection as the diversity objective. Optimising this then moves the sample in a direction diverse compared to previous samples. 
 }  
 \label{fig:placeholder}
\end{figure}

\subsection{Diffusion Language Models}
Unlike AR models which generate tokens sequentially, DLMs such as LLaDA generate all tokens in a sequence simultaneously through an iterative refinement process~\citep{nie_large_2025}.
Formally, we define a forward process that gradually corrupts a clean data sample $\mathbf{x}^0 \in \mathbb{R}^{S \times V}$ (where $S$ is sequence length and $V$ is vocabulary size) into a fully masked state $\mathbf{x}^T$ over $T$ timesteps.
This corruption is modeled as a masking process, where each token is independently replaced by a special \texttt{[MASK]} token with probability determined by a time-dependent schedule $\gamma(t)$.
Specifically, the forward transition probability for a token at sequence position $j$, denoted $x^t_j$, given the clean token $x^0_j$ is defined as:
\begin{equation}
 q(x^t_j | x^0_j) = \begin{cases} 
 1 - \gamma(t) & \text{if } x^t_j = x^0_j \\
 \gamma(t) & \text{if } x^t_j = \texttt{[MASK]} \\
 0 & \text{otherwise}
 \end{cases}
\end{equation}
where $\gamma(t)$ is a monotonically increasing function such that $\gamma(0) = 0$ and $\gamma(T) =1$.
The generative reverse process then aims to learn a denoising network $p_\theta(\mathbf{x}^{t-1} | \mathbf{x}^t)$ to recover the original tokens from their masked initial states.
Several models including LLaDA employ a re-masking strategy, where a subset of the high-confidence predictions are kept, with the rest re-masked for the next iteration \citep{nie_large_2025}. This global view of the sequence at every step captures bidirectional relationships and, in the context of this work, enables inference-time interventions on the entire remaining sequence.

\subsection{Inference-Time Diversity Sampling}
For a batch of size $B$ with a given sample $i \in \{1, \dots, B\}$, each step $t$ in the diffusion process yields predicted logits which we denote with $\mathbf{x}_i^t \in \mathbb{R}^{B \times S \times V}$.
We present a simple inference-time framework for modifying the predicted logits to penalise redundancy.
 Our method is applied independently for each time step, so we omit the $t$ and use $\mathbf{x}_i$ to refer to the $(S \times V)-$dimensional logits for sample $i$ with assumed timestep $t$.
We assume a sample $i$ with logits $\mathbf{x}_i$, previous samples $\mathbf{x}_{<i} = [\mathbf{x}_1; \cdots; \mathbf{x}_{i-1}]$, and a feature extractor $F:S\times V \to \mathbb{R}^d$.
For simplicity, we denote the feature vector for the current state as $v_i \triangleq F(\mathbf{x}_i)$, and the concatenation of feature vectors from previous states as $\mathbf{v}_{<i} \triangleq [F(\mathbf{x}_1); \cdots;F(\mathbf{x}_{i-1})]$. 
We define our diversity term as $\mathcal{L}_{\text{div}}(v_i, \mathbf{v}_{<i})$, which can be any differentiable loss function capturing diversity of features. 

For each sample $i$, our framework returns updated logits $\hat{\mathbf{x}}_i$ which are pushed to optimise the diversity term with respect to previous samples in the batch:
\begin{equation}
  \hat{\mathbf{x}}_i = \mathbf{x}_i - \alpha \cdot \nabla_{\mathbf{x}_i}\mathcal{L}_{\text{div}}(v_i, \mathbf{v}_{<i})
  \label{div_loss}
\end{equation}
where $\alpha$ is a hyperparameter controlling the strength of the repulsion.
In line with previous work~\citep{morshed_diverseflow_2025}, we anneal $\alpha$ linearly with $t$ throughout the diffusion process. Intuitively, this gives higher diversity earlier in the generation when the high level structure is being formed, and prevents excessive intervention for later steps where finer details are commonly filled in.

\subsection{Feature Extraction and Quality Scoring}
The effectiveness of a diversity intervention will depend on the choice of feature extractor $F$, which must encode the semantic or structural qualities of the generation we wish to diversify.
Several prior works have used computationally intensive features, such as pre-trained semantic encoders~\citep{shi_semantic-guided_2025} or learned similarity classifiers~\citep{li_jointly_2025}.
To ensure negligible overhead, we instead opt for a lightweight feature extraction strategy that operates directly on the model's output distribution. 

For a sample $i$ at timestep $t$, we construct a unified probability distribution $P_i \in \mathbb{R}^{S \times V}$ that captures both the model's uncertainty and its committed decisions.
For indices $s$ corresponding to currently masked tokens, we compute the softmax over the predicted logits: $P_{i,s} = \text{softmax}(\mathbf{x}_{i,s})$.
However, for unmasked tokens which have already materialised, we assign a probability of 1 to the generated token index and 0 elsewhere.
We then obtain a fixed-size representation $v_i \in \mathbb{R}^V$ by applying max-pooling over the sequence dimension $S$:
\begin{equation}
 v_i = \max_{s \in S} P_{i, s}.
\end{equation}
This vector $v_i$ represents a global confidence profile of the sequence across the vocabulary.
By setting realised tokens to 1, we ensure that the feature vector is a weighted combination of the sequence's established history and its intended future trajectory. 

\paragraph{Quality Awareness}
Strictly enforcing diversity can push the model towards low-probability, incoherent modes. To mitigate this, we also include use a scalar \textit{quality score} $q_i \in \mathbb{R}$ as part of the feature extractor.
We define $q_i$ as the average maximum confidence of the unmasked tokens in the sequence:
\begin{equation}
 q_i = \frac{1}{|U_i|} \sum_{s \in U_i} \max_{v} P_{i, s, v}
\end{equation}
where $U_i$ is the set of unmasked token indices for sample $i$.
By weighting with $q_i$, the model is discouraged from excessive diversity in areas it is confident in, preserving generation quality. 

\subsubsection{Orthogonal Diverse Diffusion}
Algorithm~\ref{alg} outlines our \sysname{} approach. 
For the $i$-th sample in the batch with feature vector $v_i$, we maintain an orthogonal basis for the history vectors $\mathbf{v}_{<i}$. 
This is done using the Gram-Schmidt algorithm, where we define $B_{<i}\triangleq[b_1, \dots, b_{i-1}]$ as the orthogonal basis for the history. 
We then set the diversity term $\mathcal{L}_{\text{div}}$ to be the negative norm of the projection of the residual of current sample features $v_i$ onto the previous samples subspace $B_{<i}$, scaled by the sample's quality score $q_i$: 
\begin{equation}
   \mathcal{L_{\text{orth}}}(v_i, \mathbf{v}_{<i})\triangleq q_i \cdot \left( -||v_i - \text{proj}_{B_{<i}}(v_i)||_2 \right).
   \label{orth_loss}
\end{equation}

The gradient 
$\nabla \mathcal{L}_{\text{orth}}$
directs the logits to minimise the magnitude of this projection, effectively forcing the model to explore the null space of previous generations.  In combination with a lightweight feature extractor $F$, this approach is extremely cheap as we are only tracking the gradients with respect to the model logits after generation. As outlined in Algorithm~\ref{alg}, we apply stop-gradients to the projection operations to treat the established subspace as a fixed target, restricting backpropagation strictly to the active token and preventing the formation of an expensive, recursive computation graph. Unlike the optimisation objective of DiverseFlow~\citep{morshed_diverseflow_2025} which optimises the batch globally, 
our method employs a greedy, sequential projection that forces each sample to independently diverge only from its predecessors. A direct consequence of this formulation is that the generation trajectory of any sample $i$ depends exclusively on the subspace spanned by earlier samples. Therefore the resulting output for sample $i$ will be batch size invariant, giving the same output across runs (given the same logits from the base model) regardless of the total batch size $k$, for any $k\ge i$.

\begin{algorithm}[t]
\SetKwInOut{Input}{Input}
\SetKwInOut{Output}{Output}
 \Input{
 Model logits $X = [\mathbf{x}_1, \dots, \mathbf{x}_n]$, 
 step size $\alpha$,
 diffusion timestep $t$,
 feature extractor $F$
 }
 \Output{Updated logits $X'$} 
 \tcp{Compute features and qualities}
 $V, Q \gets F(X)$ \tcp*{$V=[v_1, \dots, v_n]$, $Q=[q_1, \dots, q_n]$}
 
 \tcp{Initialise orthonormal basis with the first token}
 $b_1 \gets \text{sg}\left(\frac{v_1}{||v_1||_2}\right)$ \tcp*{sg() stops gradient tracking}
 
 $B_{<i} \gets [b_1]$
 
 $\alpha_t\gets (1 - \frac{1}{t})\cdot\alpha$
 
 $\mathcal{L}_{\text{total}} \gets 0$
 
 \For{$i$ in $\{2, \dots, n\}$}{
     $p_i \gets \text{proj}_{B_{<i}}(v_i)$
     
     \tcp{Accumulate loss to capture diversity with previous samples}
     
     $\mathcal{L}_{\text{orth}}^{(i)} \gets -q_i \cdot ||v_i - \text{sg}(p_i)||_2$
     
     $\mathcal{L}_{\text{total}} \gets \mathcal{L}_{\text{total}} + \mathcal{L}_{\text{orth}}^{(i)}$
    
     \tcp{Update orthonormal basis}
     $r_i \gets \text{sg}(v_i - p_i)$
     
     $b_i \gets \frac{r_i}{||r_i||_2}$ 
     
     $B_{<i} \gets [B_{<i} : b_i]$ 
}

\tcp{Update logits simultaneously}
$X' \gets X - \alpha_t \cdot \nabla_{X}\mathcal{L}_{\text{total}}$

\Return $X'$

\caption{\sysname{}: Orthogonal Diverse Diffusion}
\label{alg}
\end{algorithm}

\section{Experiments}

\begin{table*}[t]
\centering
\setlength{\tabcolsep}{8pt} 
\begin{tabular}{lccccc}
\toprule
\multirow{2}{*}{\textbf{Step Size} ($\alpha$)} & \multicolumn{5}{c}{\textbf{Temperature} ($\theta$)} \\
\cmidrule(lr){2-6}
  & 0.0 & 0.5 & 1.0 & 1.5 & 2.0 \\
\midrule
\multicolumn{6}{c}{\textsc{GSM8K (200 Problems)}} \\
\midrule
\multicolumn{6}{l}{\textit{Baseline (LLaDA)}} \\
0   & $61.0 \pm 0.0$ & $74.8 \pm 0.2$ & $81.3 \pm 0.7$ & $83.4 \pm 0.6$ & $76.5 \pm 0.4$ \\

\addlinespace
\multicolumn{6}{l}{\textit{LLaDA + DPP}} \\
2   & $63.5 \pm 0.0$ & $75.9 \pm 0.7$ & $82.3 \pm 0.7$ & $83.9 \pm 0.8$ & $77.1 \pm 0.6$ \\
8   & $62.5 \pm 0.0$ & $67.1 \pm 0.6$ & $76.6 \pm 0.2$ & $82.5 \pm 0.4$ & $79.3 \pm 0.5$ \\
16   & $61.5 \pm 0.0$ & $65.5 \pm 0.4$ & $70.2 \pm 0.5$ & $78.7 \pm 0.8$ & $80.8 \pm 0.7$ \\
32   & $61.0 \pm 0.0$ & $61.9 \pm 0.3$ & $65.2 \pm 0.9$ & $72.6 \pm 0.4$ & $78.0 \pm 0.5$ \\
64   & $64.5 \pm 0.0$ & $64.4 \pm 0.2$ & $66.8 \pm 0.4$ & $68.8 \pm 0.7$ & $77.2 \pm 0.8$ \\
128   & $66.5 \pm 0.0$ & $66.4 \pm 0.1$ & $66.8 \pm 0.2$ & $67.9 \pm 0.6$ & $76.9 \pm 0.7$ \\
\addlinespace
\multicolumn{6}{l}{\textit{LLaDA + \sysname{} }} \\
2   & $69.8 \pm 0.2$ & $80.8 \pm 0.5$ & $83.2 \pm 0.7$ & $83.1 \pm 0.5$ & $78.2 \pm 0.8$ \\
8   & $70.6 \pm 0.1$ & $81.6 \pm 0.4$ & $84.0 \pm 0.5$ & $83.4 \pm 0.3$ & $81.3 \pm 0.4$ \\
16   & $74.6 \pm 0.3$ & $82.8 \pm 0.3$ & $84.2 \pm 0.6$ & $84.4 \pm 0.3$ & $83.1 \pm 0.4$ \\
32   & $76.9 \pm 0.1$ & $85.4 \pm 0.5$ & $86.2 \pm 0.6$ & $86.4 \pm 0.5$ & $84.9 \pm 0.3$ \\
64   & $75.5 \pm 0.0$ & $87.3 \pm 0.6$ & $\mathbf{87.9 \pm 0.5}$ & $86.7 \pm 0.4$ & $85.6 \pm 0.5$ \\
128   & $\mathbf{79.0 \pm 0.4}$ & $\mathbf{87.8 \pm 0.4}$ & $87.6 \pm 0.6$ & $\mathbf{88.6 \pm 0.5}$ & $\mathbf{87.5 \pm 0.6}$ \\
\addlinespace
\midrule
\multicolumn{6}{c}{\textsc{HumanEval}} \\
\midrule
\multicolumn{6}{l}{\textit{Baseline (LLaDA)}} \\
0   & $19.5 \pm 0.0$ & $33.5 \pm 0.4$ & $42.7 \pm 0.6$ & $41.8 \pm 0.8$ & $7.7 \pm 0.6$ \\
\addlinespace
\multicolumn{6}{l}{\textit{LLaDA + DPP}} \\
2   & $31.1 \pm 0.0$ & $36.7 \pm 0.6$ & $46.6 \pm 0.4$ & $43.9 \pm 0.5$ & $7.2 \pm 0.7$ \\
8   & $32.9 \pm 0.0$ & $37.9 \pm 0.5$ & $44.2 \pm 0.6$ & $45.3 \pm 0.5$ & $13.6 \pm 0.5$ \\
16   & $32.9 \pm 0.0$ & $35.6 \pm 0.3$ & $39.4 \pm 0.6$ & $42.6 \pm 0.9$ & $26.4 \pm 1.0$ \\
32   & $29.9 \pm 0.0$ & $28.7 \pm 0.2$ & $31.1 \pm 0.6$ & $33.7 \pm 0.6$ & $33.7 \pm 0.7$ \\
64   & $19.5 \pm 0.0$ & $19.1 \pm 0.1$ & $18.7 \pm 0.3$ & $24.7 \pm 0.7$ & $28.5 \pm 0.8$ \\
128   & $12.8 \pm 0.0$ & $12.8 \pm 0.0$ & $12.3 \pm 0.3$ & $15.6 \pm 0.6$ & $21.7 \pm 1.0$ \\
\addlinespace
\multicolumn{6}{l}{\textit{LLaDA + \sysname{}}} \\
2   & $28.1 \pm 0.1$ & $38.2 \pm 0.6$ & $44.8 \pm 0.5$ & $43.4 \pm 0.4$ & $8.1 \pm 0.6$ \\
8   & $38.3 \pm 0.7$ & $47.2 \pm 0.8$ & $\mathbf{51.3 \pm 0.6}$ & $\mathbf{51.6 \pm 0.4}$ & $17.5 \pm 0.5$ \\
16   & $\mathbf{41.5 \pm 0.1}$ & $\mathbf{48.8 \pm 0.7}$ & $\mathbf{51.3 \pm 0.7}$ & $49.8 \pm 0.8$ & $33.7 \pm 1.1$ \\
32   & $41.3 \pm 0.2$ & $44.9 \pm 0.6$ & $48.6 \pm 0.7$ & $47.3 \pm 0.5$ & $\mathbf{40.2 \pm 0.7}$ \\
64   & $41.3 \pm 0.2$ & $45.6 \pm 0.6$ & $46.2 \pm 0.4$ & $45.1 \pm 0.9$ & $37.7 \pm 0.3$ \\
128   & $38.4 \pm 0.0$ & $43.7 \pm 0.7$ & $46.3 \pm 0.8$ & $44.3 \pm 0.7$ & $38.5 \pm 1.3$ \\

\addlinespace
\bottomrule
\end{tabular}
\caption{Pass@16 results (mean $\pm$ SE over 8 runs) for GSM8K and HumanEval over various temperatures ($\theta$) and step sizes ($\alpha$). \textbf{Bold} values indicate the best result for each $\theta$. LLaDA + DPP (Algorithm~\ref{alg:joint}) uses a global DPP objective~\citep{morshed_diverseflow_2025}. LLaDA + \sysname{} uses our approach in Algorithm~\ref{alg}.}
\label{tab:results_final}
\end{table*}

\subsection{Experimental Setup}
We evaluate our approach using the LLaDA-8B-Instruct model~\citep{nie_large_2025}, which we quantise to 4-bit for our experiments.
We compare three setups. The first is standard LLaDA with no logit intervention.
For the second approach, we update logits using the DPP based loss from~\citep{morshed_diverseflow_2025} as our objective $\mathcal{L}_{div}$ from Equation~\ref{div_loss} (Outlined in Algorithm~\ref{alg:joint}).
This tests our diversity framework with a global batch diversity objective, which has seen success in image diffusion.
Finally, we test our \sysname{} approach as outlined in Algorithm~\ref{alg}, where we use the loss in Equation~\ref{orth_loss} as $\mathcal{L}_{div}$.
We conduct experiments on two standard benchmarks, HumanEval~\citep{chen_evaluating_2021} and GSM8K~\citep{cobbe_training_2021}.
We take the first 200 problems in GSM8K to facilitate a large scale analysis of our approach over varying temperature and alpha levels, with 8 runs per configuration to improve confidence in the results.
We compare approaches across a grid of temperature settings, as well as $\alpha$ values for \sysname{} to investigate the influence of the step size on the final results. 
We fix the number of diffusion steps to 32, with a generative length of 64. The batch size is fixed to 16 and we report the empirical Pass@$16$, which is the percentage of outputs which found the correct answer in \textit{at least one of} the 16 attempts. 

\subsection{Results}
\label{sec:results}

\begin{table}[t]
\centering
\setlength{\tabcolsep}{10pt} 
\begin{tabular}{lcc}
\toprule
\textsc{Approach} & \textsc{GSM8K} & \textsc{HumanEval} \\
\midrule
\textit{Baseline (s)} & $22.5 \pm 0.3$  & $30.8 \pm 0.4$ \\
\textit{\sysname{} (s)} & $23.8 \pm 0.5$ & $32.0 \pm 0.2$\\
\addlinespace
\textbf{Overhead (\%)} & \textbf{+5.8\%} & \textbf{+3.9\%} \\
\bottomrule
\addlinespace
\end{tabular}

\caption{Wall time comparison for generation between standard LLaDA (Baseline) and \sysname{} (Batch Size 16). We report the execution time per batch (in seconds) averaged over all $\alpha$ and temperature settings for each approach, and the relative overhead percentage. Our method introduces minimal latency across both benchmarks, and scales independently of model size.}
\label{tab:latency}
\end{table}

Table~\ref{tab:results_final} presents our main results. We observe a consistent and significant improvement in our approach across both benchmark tasks. As expected, the improvements are particularly large for temperature $\theta=0.0$, which corresponds to the greedy decoding regime where $\text{Pass}@k = \text{Pass}@1$. For $\theta=2$, the improvements are especially noteworthy in the HumanEval benchmark. For HumanEval, the generated output must correctly compile and pass test cases, and so requires both correct syntax and reasoning for the complete output. Excessive diversity at the token level, which we observe at $\theta=2$, therefore results in significantly reduced performance for the baseline. As our approach optimises both diversity and quality, it provides a mechanism to correct for this without tuning. For GSM8K, in all cases, we observe a consistent increase in performance as we increase our step size $\alpha$. The improvements in HumanEval are significant, however we observe large $\alpha$ values decrease performance. Given the strict correctness requirements discussed above, this likely corresponds to the inflection point of the diversity-quality tradeoff. The final key observation is that our approach is far less sensitive to temperature than baseline, which gives confidence that our approach provides improvements with minimal tuning.

Interestingly, while the DPP objective from DiverseFlow~\citep{morshed_diverseflow_2025} globally optimises
across all samples, our experiments demonstrate that our greedy, stop-gradient approach results in improved Pass@$k$. One possible explanation is that since joint optimisation causes all samples to repel each other, this may push better, high probability samples off the optimal mode to achieve diversity. Another possible disadvantage is the complex dependencies between all samples in a batch, which may create chaotic optimisation trajectories when optimised jointly. In contrast, our objective is simply the projection onto the detached subspace of previous samples, which is a static hyperplane. 

Standard sampling treats every generation independently, with no shared information between attempts. Consequently, increasing the batch size often results in wasted compute, as the model repeatedly samples the most likely modes.
In contrast, \sysname{} introduces a batch-aware dependency that explicitly trades off exploitation (quality) and exploration (diversity). 
While individual sample quality (Pass@1) can slightly decrease as we force the model to explore less probable paths (see Figure~\ref{fig:tradeoff}), the coverage of the solution space (Pass@16) increases significantly. 
This confirms that our method effectively converts compute into useful exploration, ensuring that each additional sample contributes a unique perspective to the batch rather than repeating previous failure modes.

\subsubsection{Overhead Analysis}

Table~\ref{tab:latency} shows the average time overhead between ODD and LLaDA with standard sampling, for the experiments in Section~\ref{sec:results}. We observe only a small increase for both benchmarks, particularly considering the rate of improvement in Pass@$16$. As our approach only optimises logits after they are computed, the overhead is independent of the base model, and so will have better relative overhead as model size increases. Furthermore, the operations used by \sysname{} are lightweight in comparison to the model forward pass.
We conduct a more detailed overhead analysis in Appendix~\ref{overhead}.

\section{Conclusion and Future Work}

We presented \sysname{}, a training free inference intervention that significantly enhances the diversity and Pass@$k$ performance of Diffusion Language Models. 
By sequentially projecting intermediate logits away from the subspace of previous samples, we effectively minimise sampling redundancy at negligible computational cost. 
Our results on HumanEval and GSM8K demonstrate that this simple geometric repulsion enables more effective exploration of the solution space, particularly in domains requiring complex reasoning where valid solutions are sparse.

Our work highlights a unique advantage of the diffusion paradigm, which is the ability to intervene globally on the generation process. Unlike Autoregressive models, diffusion models allow for global sequence optimisation, enabling low cost improvements to sample efficiency. With inference time compute becoming a major factor in scaling reasoning capabilities, our approach helps improve the efficiency of resources during generation.

\subsection{Future Work}
Future work should explore the effectiveness of different feature extractors. There are possible improvements to be made by considering, for example, the positional information in tokens. The semantics of partial sequences could be captured, as has been done in AR models (e.g. \cite{shi_semantic-guided_2025, li_jointly_2025}), although the overhead could be a limiting factor. Dimensionality reduction of the feature space could also be explored to further reduce the overhead of our approach. 

\bibliographystyle{abbrv}
\bibliography{references}

\clearpage

\appendix
\section{Cumulative Results}
\begin{table}[t]
\centering
\begin{tabular}{@{}llcc@{}}
\toprule
\textbf{Dataset} & \textbf{$\alpha$ (Repulsion)} & \textbf{Solved (Union)} & \textbf{Coverage} \\ \midrule
\multirow{7}{*}{\textbf{GSM8K}}     & 0.0 (Baseline)  & \textbf{198}           & \textbf{99.00\%} \\ 
                                    & 2.0             & 193                    & 96.50\%         \\ 
                                    & 8.0             & 192                    & 96.00\%         \\ 
                                    & 16.0            & 195                    & 97.50\%         \\ 
                                    & 32.0            & 195                    & 97.50\%         \\ 
                                    & 64.0            & 195                    & 97.50\%         \\ 
                                    & 128.0           & 197                    & 98.50\%         \\ 
\midrule
\multirow{7}{*}{\textbf{HumanEval}} & 0.0 (Baseline)  & 110                    & 67.07\%         \\ 
                                    & 2.0             & 113                    & 68.90\%         \\ 
                                    & 8.0             & 124                    & 75.61\%         \\ 
                                    & 16.0            & \textbf{129}           & \textbf{78.66\%} \\ 
                                    & 32.0            & 122                    & 74.39\%         \\ 
                                    & 64.0            & 123                    & 75.00\%         \\ 
                                    & 128.0           & 120                    & 73.17\%         \\ 
\bottomrule
\end{tabular}
\vspace{1ex}
\caption{Cumulative problem coverage across all temperature settings. \sysname{} ($\alpha > 0$) significantly improves coverage on HumanEval, finding many problems missed by the baseline.}
\label{cumulative}
\end{table}

Table~\ref{cumulative} presents the cumulative problem coverage, calculated as the union of all successfully solved problems across all runs in Section~\ref{sec:results} for each temperature regime (5 temperature settings with 8 runs per benchmark, for Pass@16 gives us 640 samples per problem). Because \sysname{} generates dependent samples, this metric better reflects the bounds of the reachable solution space rather than standard independent Pass@k scaling, which we observe in Figure~\ref{fig:passk}.
On HumanEval, where strict syntactic and logical constraints make solutions sparse, baseline sampling covers only 110 problems (67.1\%).
\sysname{} ($\alpha=16$) expands this to 129 problems (78.7\%), representing a significant 17.3\% relative improvement.
Even for the cases where $\alpha > 16$, there remains a significant improvement in cumulative problems solved for these suboptimal $\alpha$, where Pass@16 performance significantly drops. 
This demonstrates \sysname{}'s strength as an exploration mechanism even for poor choices of $\alpha$, where it is able to discover valid, rare reasoning paths that standard sampling misses entirely, even after 640 trials over various temperatures.

 In contrast, the baseline effectively saturates the 200-problem GSM8K subset (198 problems, 99.0\% coverage). Unlike HumanEval's strict execution constraints, GSM8K evaluates only the final numerical answer. Given a large compute budget across high temperatures, standard sampling acts as an unconstrained random walk which can occasionally find the correct final answer regardless of reasoning. \sysname{} instead enforces structured exploration over distinct generation trajectories. While this slightly reduces the absolute coverage (192-197 problems), it dramatically streamlines the search. Consequently, \sysname{} significantly improves practical sample efficiency (Table \ref{tab:results_final}), raising Pass@16 to reliably find more solutions within a more practical inference budget.

\section{Diversity Dynamics}
\FloatBarrier
\begin{figure}[htbp]
 \centering
 \begin{minipage}[b]{0.79\textwidth}
  \centering
  \includegraphics[width=\linewidth]{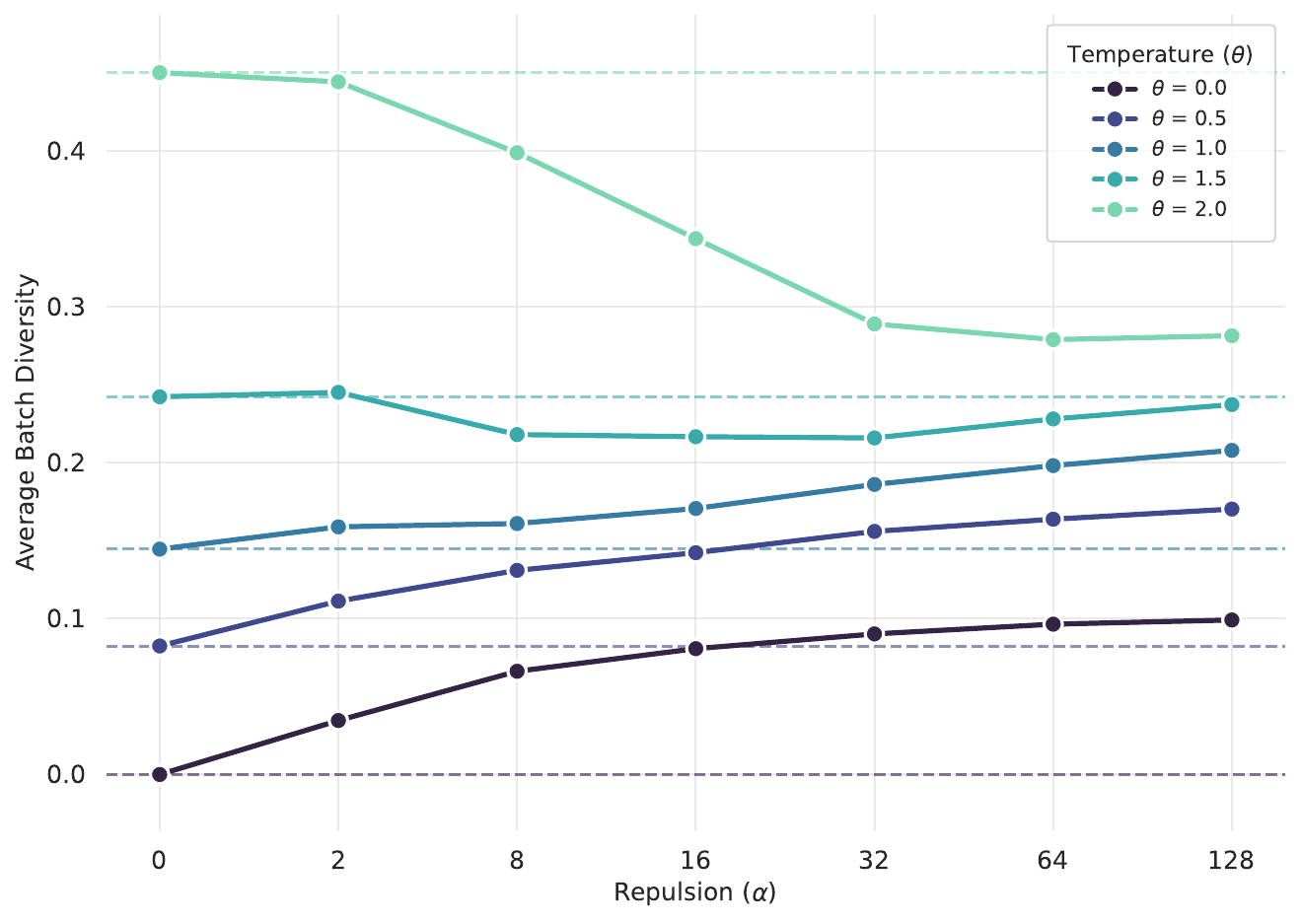}
 \end{minipage}
 \hfill
 \begin{minipage}[b]{0.79\textwidth}
  \centering
  \includegraphics[width=\linewidth]{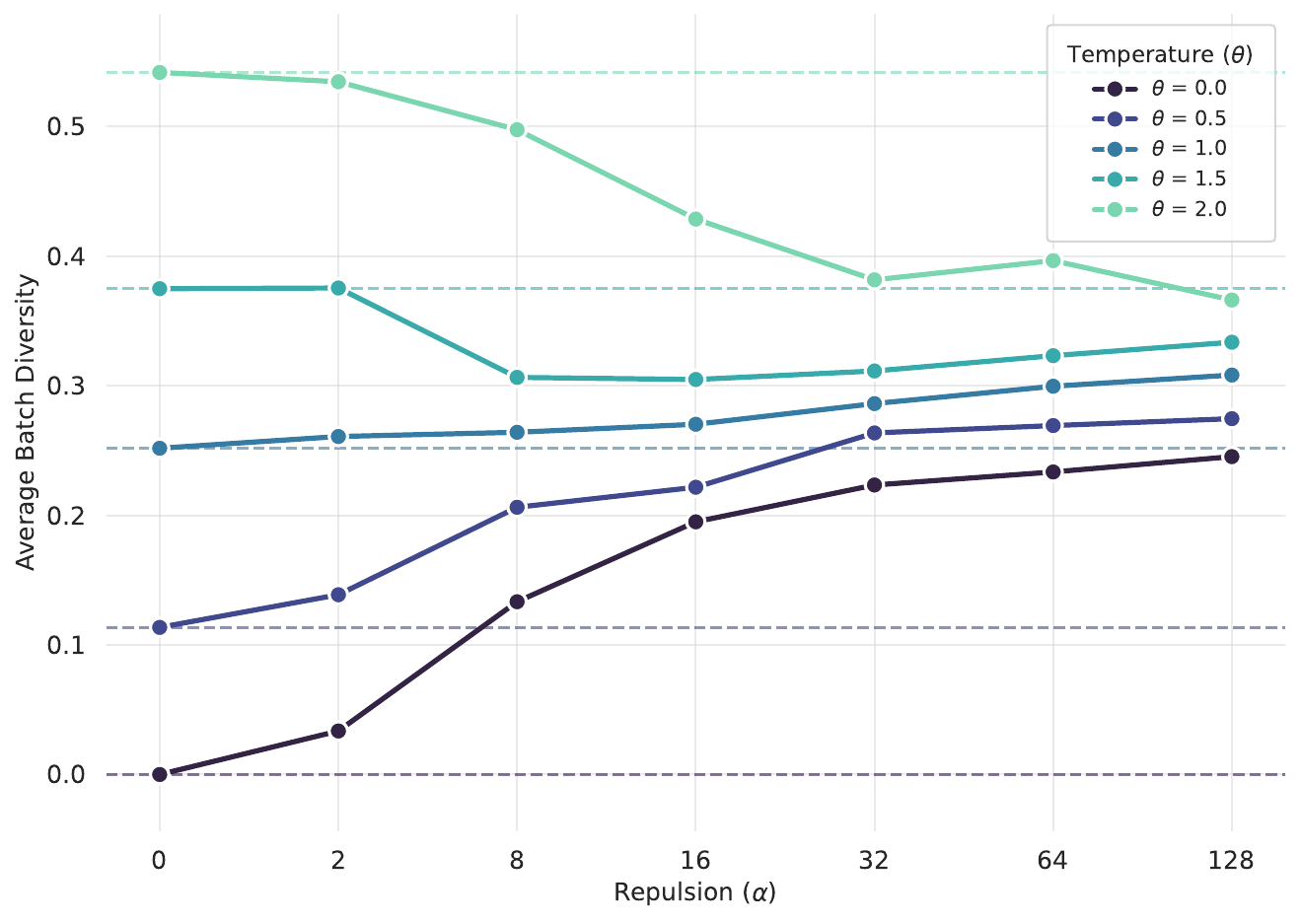}
 \end{minipage}

 \caption{Average batch diversity (1 - cosine similarity of sentence embeddings) for GSM8K (top) and HumanEval (bottom), comparing \sysname{} to the baseline ($\alpha=0$, as indicated by dashed lines). \sysname{} increases diversity at low temperatures and acts as a coherence filter at high temperatures (evidenced by the improved Pass@16 in Table~\ref{tab:results_final}), effectively balancing exploration and quality.}
 \label{fig:div}
 \end{figure}

In Figure~\ref{fig:div}, we analyse the semantic diversity of the generated batches using the average cosine similarity between sentence embeddings.  
We observe an interesting dynamic behavior in \sysname{} depending on the temperature regime:
\begin{itemize}
    \item \textbf{Low Temperature ($\theta \le 1.0$):} The baseline exhibits high similarity, with mode collapse in lower temperatures. \sysname{} significantly reduces similarity, forcing the model to explore diverse paths even when the underlying probability distribution is sharp.
    \item \textbf{High Temperature ($\theta \ge 1.5$):} The baseline naturally produces diverse but often incoherent outputs (lower similarity). Interestingly, \sysname{} tends to \textit{reduce} diversity compared to the baseline in this regime. This suggests our quality-weighted repulsion acts as a filter, guiding the model back towards high-confidence regions when the noise level is excessive.
\end{itemize}
This demonstrates that \sysname{} naturally trades off diversity and quality, by adaptively promoting diversity when needed and also promoting coherence when diversity is already sufficient.

\section{Pareto Efficiency and Trade-offs}
\begin{figure}[htbp]
 \centering
 \begin{minipage}[b]{0.79\textwidth}
  \centering
  \includegraphics[width=\linewidth]{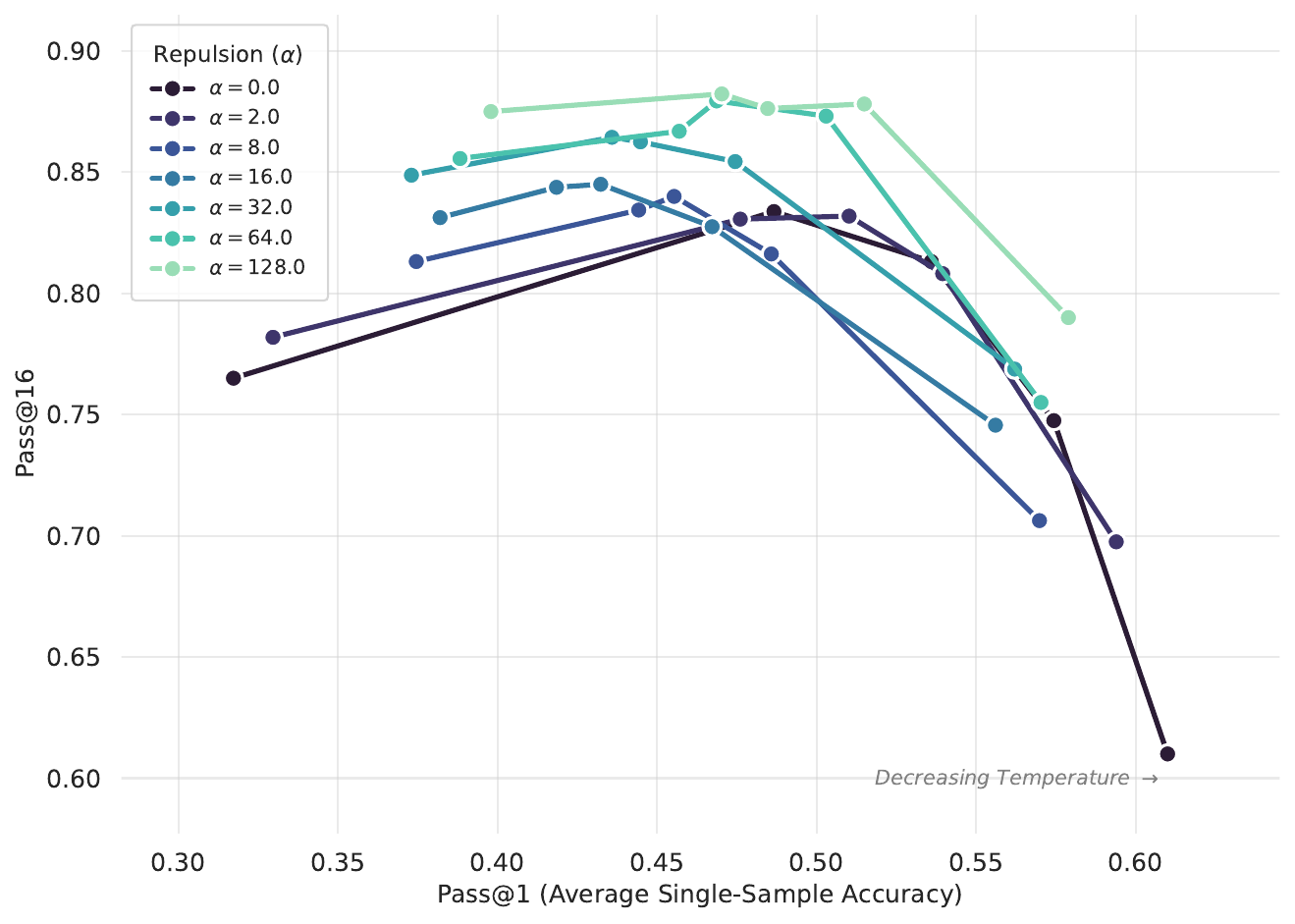}
 \end{minipage}
 \hfill
 \begin{minipage}[b]{0.79\textwidth}
  \centering
  \includegraphics[width=\linewidth]{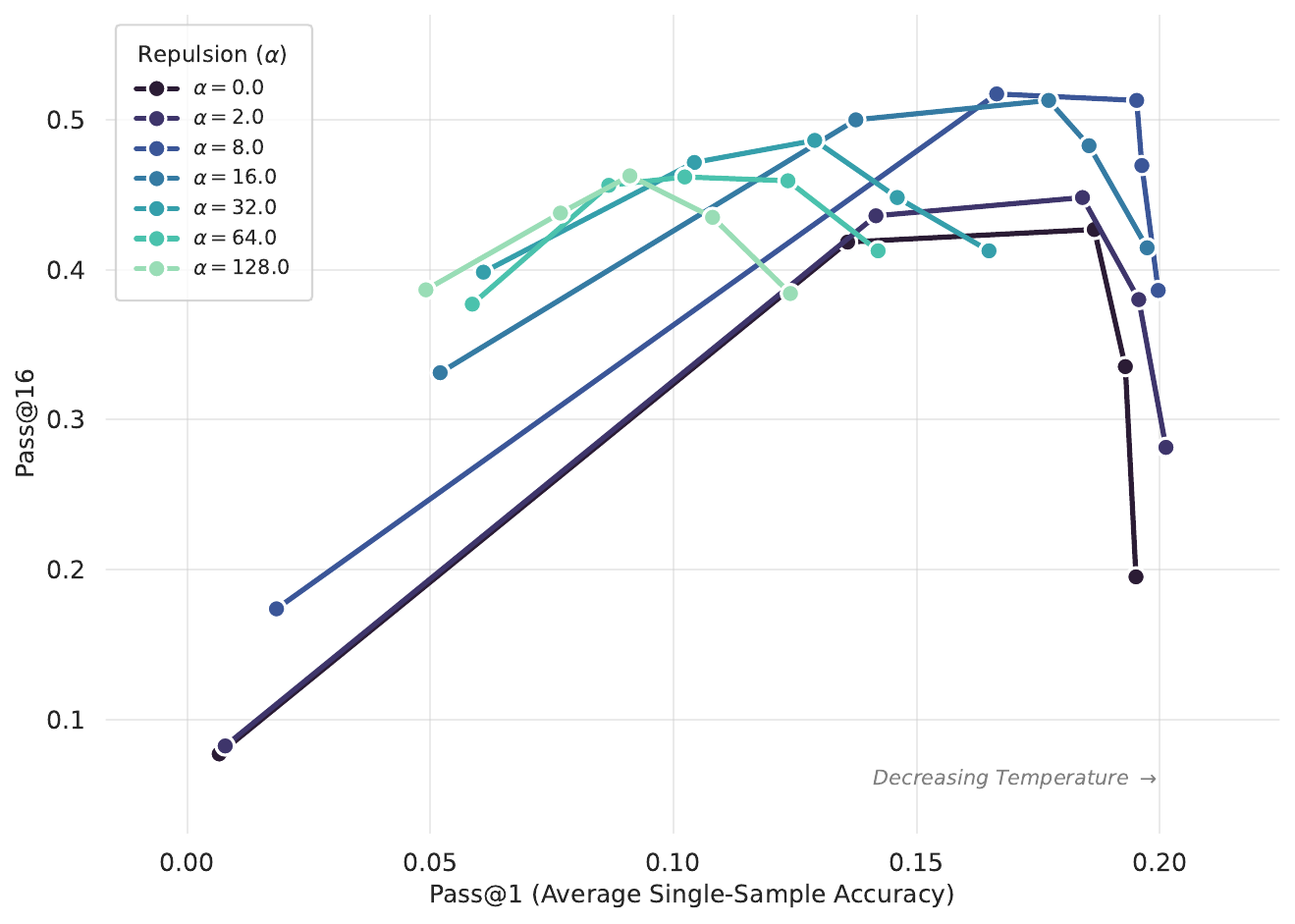}
 \end{minipage}
 
 \caption{Pass@1 vs Pass@16 Pareto frontiers. $\alpha=0$ represents standard LLaDA sampling (baseline). Data points represent decreasing temperatures from right ($\theta=0$) to left ($\theta=2$). 
 \textbf{Top (GSM8K):} \sysname{} trades off individual accuracy (shifting left) for superior batch coverage (shifting up).
 \textbf{Bottom (HumanEval):} For $\alpha \le 16$, \sysname{} achieves a Pareto improvement, boosting coverage without loss of quality.}
 \label{fig:tradeoff}
\end{figure}

Figure~\ref{fig:tradeoff} visualises the efficiency of our sampling strategy by plotting individual sample quality (Pass@1, as determined by the average correctness in a batch) against batch coverage (Pass@16).
\begin{itemize}
    \item \textbf{HumanEval (Pareto Improvement):} For code generation, moderate repulsion ($\alpha \le 16$) yields a vertical lift over the baseline. We achieve significantly higher coverage (Pass@16) without degrading the quality of individual samples (Pass@1). This indicates that \sysname{} provides a Pareto improvement over baseline for lower alpha values.
    \item \textbf{GSM8K (Exploration Trade-off):} For reasoning tasks, we observe a shift up and to the left. To maximise batch coverage (Pass@16), \sysname{} sacrifices some individual sample accuracy (Pass@1). This confirms that the model is being forced to explore less probable and riskier reasoning paths. However, the net result is a higher probability of finding the correct answer within a single batch, a capability the baseline cannot match even at high temperatures.
\end{itemize}

\section{Empirical Pass@$k$}
\begin{figure}[htbp]
 \centering
 \begin{minipage}[b]{0.89\textwidth}
  \centering
  \includegraphics[width=\linewidth]{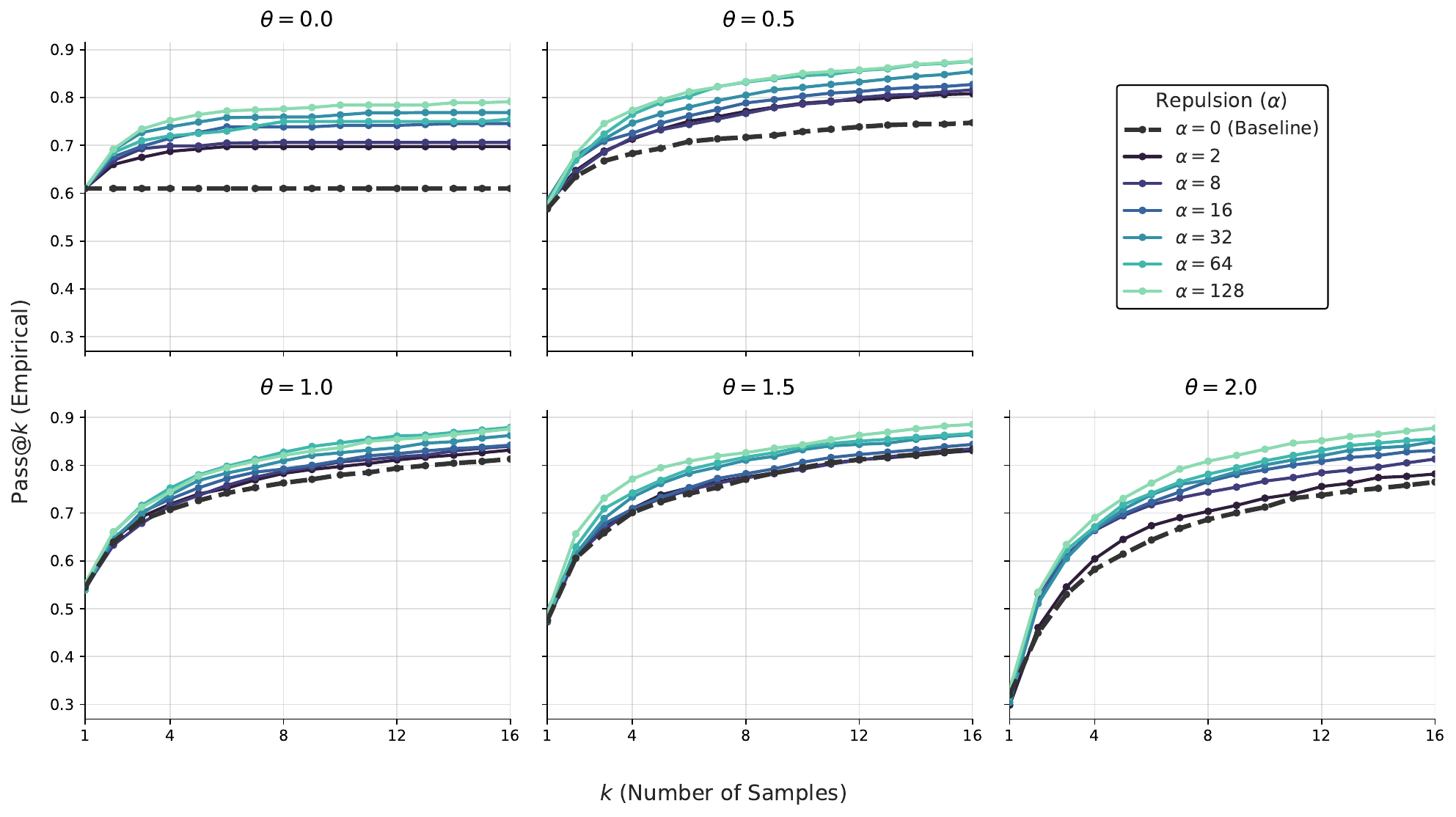}
 \end{minipage}
 \hfill
 \begin{minipage}[b]{0.89\textwidth}
  \centering
  \includegraphics[width=\linewidth]{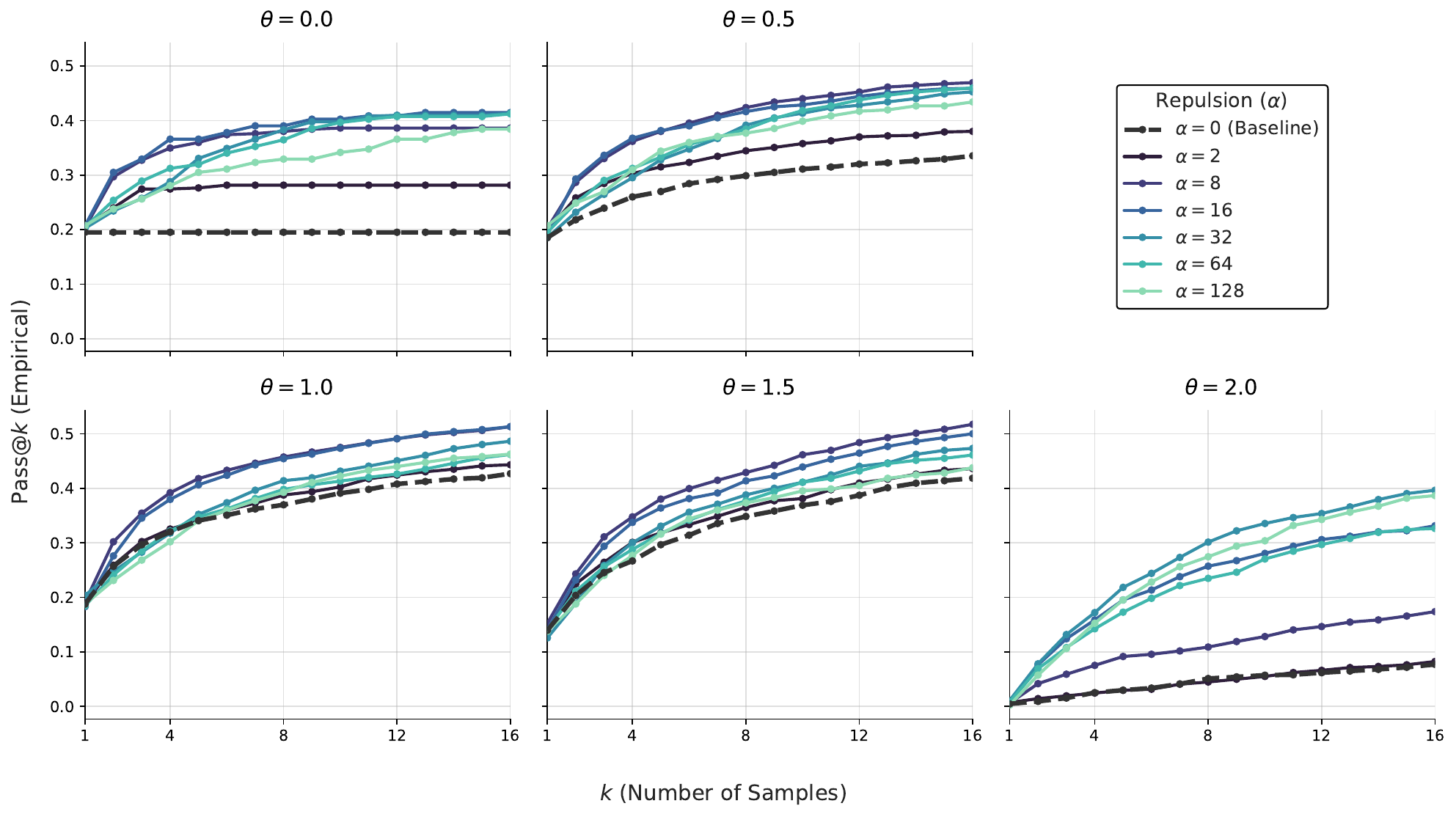}
 \end{minipage}
 
 \caption{Empirical Pass@$k$ for \sysname{} over GSM8K (top) and HumanEval (bottom). We observe a consistent improvement relative to baseline for \sysname{}. Given the batch size invariance of \sysname{}, we expect this trend would continue for higher batch sizes.}
 \label{fig:passk}
\end{figure}
Figure~\ref{fig:passk} illustrates the empirical Pass@$k$ scaling behavior of \sysname{} compared to standard baseline sampling across both the GSM8K and HumanEval benchmarks. Empirical Pass@$k$ measures the probability that at least one correct solution is found within a generated subset of $k$ samples, where we take the first $k$ out of the 16 samples from our experiments in Section~\ref{sec:results}.

As observed in the baseline curves, standard independent sampling suffers from severe diminishing returns as $k$ scales. Because the samples are generated independently, the model frequently collapses into identical failure modes, causing the curves to plateau early. In contrast, \sysname{} maintains a significantly steeper trajectory as $k$ increases. By explicitly penalising redundancy and forcing the model into orthogonal subspaces, \sysname{} ensures that every additional unit of compute (each subsequent sample up to $k=16$) explores a genuinely novel reasoning path or syntactic structure. 

Given the batch size invariance of \sysname{}, the curves shown for $k \le 16$ are identical to the prefix of the curve for any larger batch size $K > 16$. This predictable scaling property suggests that \sysname{} will continue to efficiently expand the explored solution space as compute budgets increase.

\section{DiverseFlow Baseline}
\begin{algorithm}[htbp]
\SetKwInOut{Input}{Input}
\SetKwInOut{Output}{Output}
 \Input{
 Model logits $X = [\mathbf{x}_1, \dots, \mathbf{x}_n]$, 
 step size $\alpha$,
 diffusion timestep $t$,
 feature extractor $F$,
 numerical jitter $\epsilon$
 }
 \Output{Updated logits $X'$} 
 \tcp{Compute normalised features and qualities}
 
 $V, Q \gets F(X)$ \tcp*{$V \in \mathbb{R}^{n \times d}$, $Q \in \mathbb{R}^n$}
 
 $\alpha_t\gets (1 - \frac{1}{t})\cdot\alpha$
 
 \tcp{Construct the DPP L-ensemble matrix}
 $K \gets V V^\top$ \tcp*{Feature similarity kernel}
 
 $Q_{\text{mat}} \gets Q Q^\top$ \tcp*{Quality outer product, $Q_{\text{mat}}[i,j] = q_i q_j$}
 
 $L \gets K \odot Q_{\text{mat}}$ \tcp*{Element-wise scaling}
 
 \tcp{Compute the negative log-likelihood of diversity}
 $\mathcal{L}_{\text{joint}} \gets - \Big( \log\det(L + \epsilon I) - \log\det\big(L + (1+\epsilon)I\big) \Big)$

 \tcp{Update all logits simultaneously via full backprop}
 $X' \gets X - \alpha_t \cdot \nabla_{X}\mathcal{L}_{\text{DPP}}$

\Return $X'$

\caption{Joint Batch Optimisation (DiverseFlow~\citep{morshed_diverseflow_2025})}
\label{alg:joint}
\end{algorithm}

To evaluate our approach against previous global diversity methods, we adapt the global optimisation strategy of DiverseFlow~\citep{morshed_diverseflow_2025}, originally proposed for continuous image diffusion. The methodology is detailed in Algorithm~\ref{alg:joint}.

This baseline models the diversity of the generated batch using a Determinantal Point Process~\citep{kulesza_determinantal_2012} kernel. Given the extracted feature vectors $V$ and quality scores $Q$ derived from the model logits, the algorithm constructs an L-ensemble Gram matrix $L$. This matrix is formed via the element-wise multiplication of the pure feature similarity kernel ($K = VV^\top$) and the quality outer product ($Q_{\text{mat}}$), weighting the pairwise similarities by the samples respective confidence scores. The objective $\mathcal{L}_{\text{DPP}}$ then minimises the negative log-determinant of $L$ (stabilised by a numerical jitter $\epsilon$), which geometrically corresponds to maximising the $n$-dimensional volume enclosed by the batch feature vectors.

\label{overhead}
\begin{figure}[htbp]
 \centering
 \includegraphics[width=\linewidth]{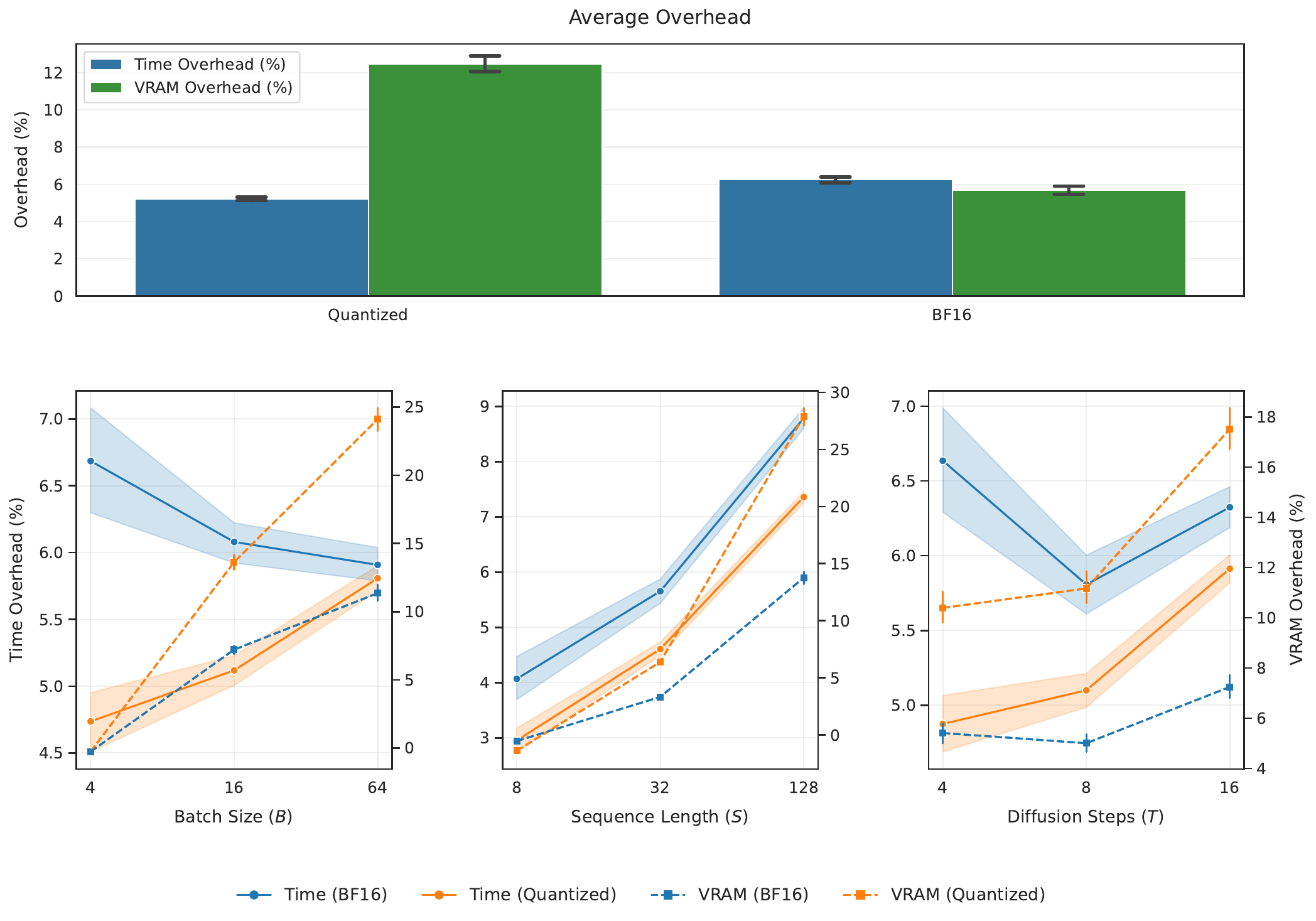} 
 \caption{Computational overhead of \sysname{} compared to standard sampling. \textbf{Top:} Global averages across all runs. \textbf{Bottom:} Scaling behavior demonstrating time and memory overheads across batch size ($B$), sequence length ($S$), and diffusion steps ($T$). \sysname{} incurs less than a $10\%$ relative time penalty, with memory scaling independently of the base model size.}
 \label{fig:overhead}
\end{figure}

\FloatBarrier
\section{Computational Overhead}

To comprehensively evaluate the computational overhead of \sysname{}, we profiled generation using the LLaDA-8B-Instruct model in both standard BF16 and 4-bit quantised precision. We measured latency and VRAM footprint scaling across batch size ($B$), generated sequence length ($S$), and diffusion steps ($T$), with 24 configurations for each problem in HumanEval and the 200 problem subset of GSM8K. The results are presented in Figure~\ref{fig:overhead}. 
Overall, \sysname{} incurs minimal latency, effectively acting as a low cost intervention for scaling $\text{Pass}@k$. Across all configurations, the time overhead consistently remains below $10\%$.
Importantly, the absolute memory and compute introduced by \sysname{} are entirely independent of the base model.
Although our approach is model agnostic, the relative latency penalty is slightly higher for the BF16 model. This corresponds to the speed decrease from quantisation overheads, with the base model being approximately 2.4s slower in the quantised setting. As expected, the relative memory overhead for the larger BF16 model is below that of the quantised model. 

We observe distinct scaling dynamics across the generation parameters. Because \sysname{} sequentially tracks the history of the batch, its memory overhead scales at $O(B^2)$ with batch size ($B$). The VRAM penalty remains negligible ($5-15\%$) while time overhead remains flat. 
Scaling the generated sequence length ($S$) results in a linear increase in both time and memory overhead. This is driven by the requirements of maintaining and backpropagating through the $B \times S \times V$ graph during the feature extraction phase. Despite this linear scaling, the peak overhead at $S=128$ remains highly manageable. It could be further alleviated by taking a subset of logits (e.g. top-$k$) rather than the whole vocabulary for each token. 
Finally, the relative time and memory overheads are invariant to the number of diffusion steps ($T$). Because the intervention is applied locally at each denoising timestep, its computational cost scales in proportion to the forward pass of the base model. 

\end{document}